\documentclass[10pt,twocolumn,letterpaper]{article}
\usepackage[pagenumbers]{cvpr} 

\usepackage[dvipsnames]{xcolor}

\newcommand{\minisection}[1]{\noindent{\textbf{#1}.}}
\usepackage{multirow}

\newcommand{\bb}{\boldsymbol{b}}
\usepackage{colortbl}
\usepackage{color}
\usepackage[utf8]{inputenc}
\newcommand{\tablestyle}[2]{\setlength{\tabcolsep}{#1}\renewcommand{\arraystretch}{#2}\centering\footnotesize}
\newlength\savewidth\newcommand\shline{\noalign{\global\savewidth\arrayrulewidth
  \global\arrayrulewidth 1pt}\hline\noalign{\global\arrayrulewidth\savewidth}}
\usepackage{url}
\usepackage{booktabs}
\usepackage{amssymb}
\usepackage{bbding}
\usepackage{pifont}
\usepackage{wasysym}
\usepackage{utfsym}
\usepackage{fontawesome}
 
\newcommand{\Ours}{ODAPT\xspace}

\newcommand{\amir}[1]{{\color{red}\textbf{[AB:} #1]}}

\definecolor{cvprblue}{rgb}{0.21,0.49,0.74}
\usepackage[pagebackref,breaklinks,colorlinks,citecolor=cvprblue]{hyperref}

\def\paperID{4178} 
\def\confName{CVPR}
\def\confYear{2024}

\title{Object-based (yet Class-agnostic) Video Domain Adaptation}

\author{Dantong Niu\textsuperscript{1} \qquad Amir Bar\textsuperscript{1,2} \qquad Roei Herzig\textsuperscript{1,2} \qquad Trevor Darrell\textsuperscript{1} \qquad Anna Rohrbach\textsuperscript{1}\\  \\
\textsuperscript{1}UC Berkeley   \qquad \textsuperscript{2}Tel Aviv University}

\def\Pipeline#1{
\begin{figure*}[#1]
\begin{center}
\includegraphics[width=\linewidth]{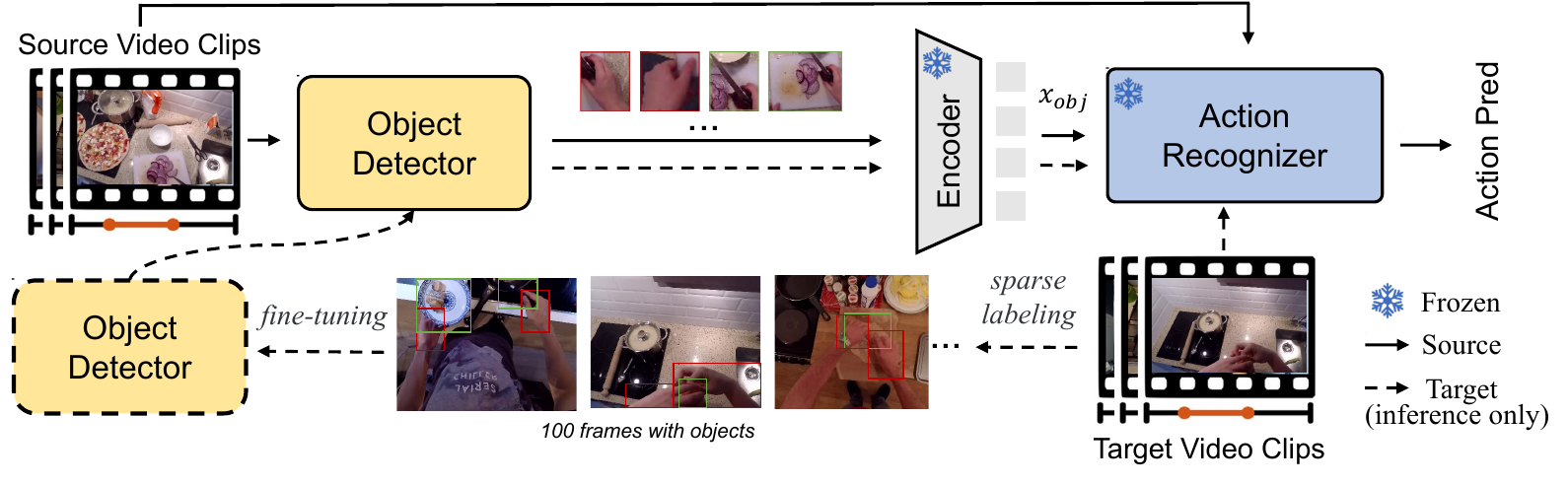}
\caption{\textbf{The overview of the \Ours.} (Top) we illustrate a generic action recognition model trained in the source domain that utilizes an object detector to predict objects as prior knowledge. (Bottom  with dotted line) we illustrate the adaptation process via fine-tuning the object detector alone based on a set of sparse labeled bounding boxes from the target domain, keeping the rest structure frozen. \textbf{In the adaptation, the object detector is replaced with the adapted one}.}
\end{center}
\vspace{-0.8em}
\label{fig: overview}
\end{figure*}
}

\begin{document}
\maketitle

\begin{abstract}

Existing video-based action recognition systems typically require dense annotation and struggle in environments when there is significant distribution shift relative to the training data. Current methods for video domain adaptation typically fine-tune the model using fully annotated data on a subset of target domain data or align the representation of the two domains using bootstrapping or adversarial learning.  
Inspired by the pivotal role of objects in recent supervised object-centric action recognition models, we present Object-based (yet Class-agnostic) Video Domain Adaptation (\Ours), 
a simple yet effective framework for adapting the existing action recognition systems to new domains by utilizing a sparse set of frames with class-agnostic object annotations in a target domain. Our model achieves a +6.5 increase when adapting across kitchens in Epic-Kitchens and a +3.1 increase adapting between Epic-Kitchens and the EGTEA dataset. \Ours  is a general framework that can also be combined with previous unsupervised methods, offering a  +5.0 boost when combined with the self-supervised multi-modal method MMSADA~\cite{munro2020multi} and a +1.7 boost when added to the adversarial-based method TA$^3$N~\cite{ta3n} on Epic-Kitchens. 
\end{abstract}    
\def\FigTeaser#1{
\begin{figure}[#1]
\begin{center}
\includegraphics[width=\linewidth]{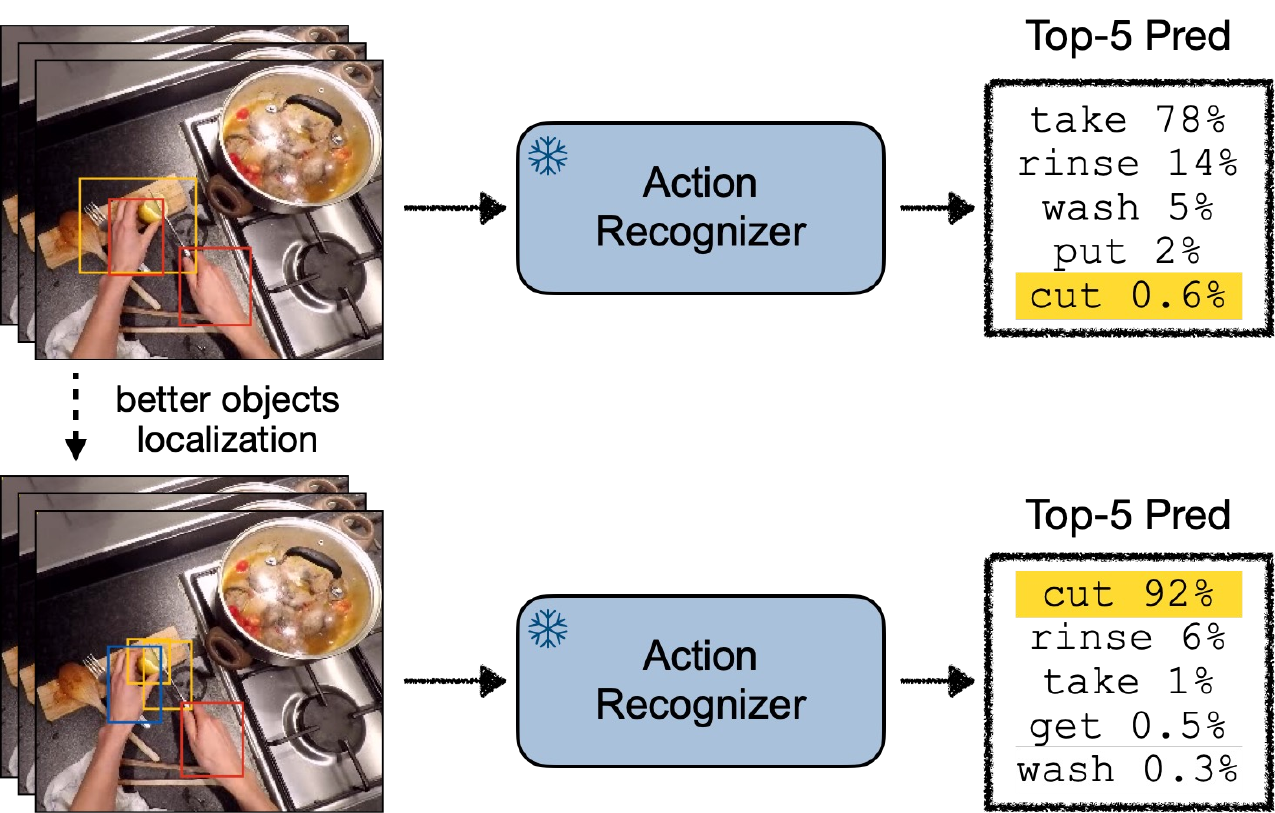}
\end{center}
\vspace{-1.2em}
\caption{\textbf{``Cut Lemon''.} We show an example which illustrates how crucial object localization is in bridging the domain gap for action recognition tasks: enhanced precision in localizing objects allows the action recognition system to more accurately understand the interactions between humans and objects, thereby improving its predictive accuracy. For instance, the system can discern specifics, such as \emph{the left hand holding a lemon and the right hand holding a knife}, indicating the \emph{cutting} action.
}
\label{fig:teaser}
\vspace{-1em}
\end{figure}
}

\FigTeaser{!t}

\section{Introduction}
\label{sec:intro}

Action or activity recognition 
performance often degrades in new domains due to the variability of objects in the new environment: \eg, different utensils may be present in different kitchens, contributing significantly to the  ``domain gap'' that reduces performance. 
%
%
%
{{Supervised domain adaptation approaches}} unreasonably assume the availability of detailed annotation in the target domain, hence interest in
{{unsupervised domain adaptation methods}} including adversarial and contrastive learning strategies~\cite{adda, dann, ta3n, choi2020shuffle, sahoo2021contrast, munro2020multi, agarwal2020unsupervised}. These methods bootstrap and/or align representations of the source and target domain in both temporal and spatial dimensions but are often computationally expensive or not adequately performant. 
%
In this work we seek a middle ground between these perspectives, and demonstrate how very sparse and class-agnostic object annotations can significantly improve video action recognition.

We present \textit{\textbf{O}}bject-based 
Video Domain A\textit{\textbf{dapt}}ation (\textbf{\Ours}), which incorporates a few key object box outlines---but not knowledge of their class labels---to improve performance in a target domain (see Fig.~\ref{fig:teaser}).  This offers an efficient compromise between purely labeled or unlabled domain adaptation approaches.   The inspiration for \Ours comes from several pioneering object-centric action recognition studies~\cite{arnab2021unified,svit2022,herzig2022orvit,Wang_videogcnECCV2018}, which have shown that the explicit modeling of objects and hands significantly enhances the recognition of complex actions, such as human-object interactions. These studies typically employ integrated object detectors to gather object and hand information to bolster the primary task. \Ours leverages this concept by utilizing a sparse set of frames from the target domain, annotated with class-agnostic object boxes, to adapt only the object-centric part of the model.  We significanly simplify the  domain adaptation process by adapting the object-centric representation and leaving the rest of the model frozen.  
Moreover, the sparse set of frames used in the adaptation are also ``object-sparse'': we only focus on hands and the objects they interact with. In practical, obtaining such data could be largely facilitated by the utilization of fundamental segmentation models such as SAM~\cite{kirillov2023segment}.


We construct two challenging scenarios to test \Ours: adaptation across 
 kitchens within \textit{Epic-Kitchens} dataset, and
adaptation 
across datasets, transferring from \textit{Epic-Kitchens} to \textit{EGTEA}. 
\textit{Epic-Kitchens}~\cite{damen2020epic} and \textit{\textit{EGTEA} Gaze+}~\cite{li2018eye} datasets are both ego-centric cooking video datasets, but distinct from each other in terms of the used hardware, recording setup, and annotation protocols, thus presenting a substantial domain gap. 

Our results show that adapting object-centric representation makes a difference in mitigating the domain gap. For a ViT-based action recognition model, \Ours achieves \textbf{+6.5} performance on the ``within dataset'' setting and \textbf{+3.1} on the ``across dataset'' setting. For CNN-based recognition models, we consider prior video domain adaptation methods~\cite{ta3n, munro2020multi,sahoo2021contrast}. When combined with these methods, \Ours brings a further boost on the ``within dataset'' setting : TA$^3$N-\Ours achieves \textbf{+1.7} increase over TA$^3$N~\cite{ta3n}; MMSADA-\Ours achieves \textbf{+5.0} increase over MMSADA~\cite{munro2020multi}; CoMix-\Ours achieves \textbf{+0.7} increase over CoMix~\cite{sahoo2021contrast}.

\noindent \textbf{Contributions.} 
\Ours is distinguished by its {{simplicity and efficacy}}, demonstrating considerable improvements in video domain adaptation with only a sparse set of target boxes. \Ours's {{versatile design}} allows it to be applied effectively across a variety of action recognition models, regardless of their structural differences. We believe \Ours is the first model to employ sparse category-agnostic box annotations for video domain adaptation, and that performance improvements with our method reflect a new baseline on these tasks. 


\def\Pipeline#1{
\begin{figure*}[#1]
\begin{center}
\includegraphics[width=\linewidth]{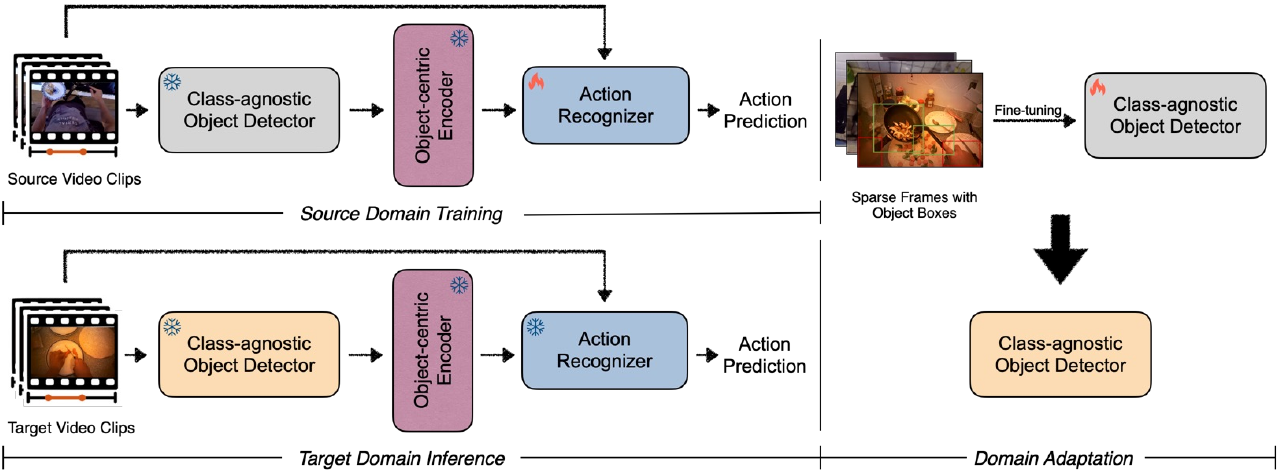}
\end{center}
\vspace{-1.2em}
\caption{\textbf{Overview of \Ours.} \Ours encompasses 
three key stages: source domain training, domain adaptation and target domain inference. Initially, the action recognizer is trained on annotated video clips in the source domain. During the adaptation phase, \Ours employs a sparse set of frames with object bounding boxes from the target domain to fine-tune the class-agnostic object detector (indicated by a ``fire''). In the target domain inference, new video clips are fed into the fine-tuned class-agnostic detector (with yellow color) and the unchanged (denoted by a ``snowflake'') action recognizer to generate predictions.}

\label{fig: overview}
\end{figure*}
}

\section{Related Work}

\noindent \textbf{Video-based Domain Adaptation.} The field of video-based domain adaptation has garnered significant attention recently. Earlier research predominantly concentrated on view-invariant action recognition, with notable studies in this area including works~\cite{kong2017deeply, li2018unsupervised,liu2017enhanced, rahmani2015learning,sigurdsson2018actor}. On the other hand, Unsupervised Domain Adaptation (UDA), particularly aimed at addressing environmental variations, has begun to attract interest more recently. Key contributions in this domain include studies~\cite{chen2019temporal, jamal2018deep, pan2020adversarial, munro2020multi, agarwal2020unsupervised}. These studies primarily achieve adaptation by minimizing/aligning feature representations between source and target domains. However, their alignment focuses on general features. In contrast, our approach, \Ours, prioritizes object-centric features. We posit that object-centric regions contain more semantically meaningful information, unique to source and target domains. Therefore, aligning these representations is likely to significantly enhance domain adaptation.

Inspired by foundational image-based domain adaptation techniques~\cite{adda, dann, chen2018domain}, most unsupervised video domain adaptation methods have adopted adversarial learning frameworks to synchronize representations across source and target domains. For instance, \cite{jamal2018deep} emphasized aligning spatial representations between domains. Similarly, \cite{chen2019temporal, pan2020adversarial} proposed methods to align temporal dynamics via learned attention mechanisms. Extending this approach, \cite{munro2020multi} applied adversarial training to multi-modal architectures, enhancing performance.

\noindent \textbf{Instance-level Domain Adaptation.} Noteworthy studies such as \cite{agarwal2020unsupervised, chen2018domain} have made strides in adapting features at both the image/frame and instance levels, utilizing domain classifier networks for this purpose. However, these works continue to rely on adversarial feature alignment which often results in protracted adaptation times. In scenarios characterized by substantial domain shifts, such as the emergence of unseen objects, the effectiveness of their RPN in accurately capturing instance-level features comes into question. This limitation is critical, as inaccurate instance-level feature capture could render the alignment process less effective or even redundant in bridging the domain gap.
Contrary to these methods, our work explores a novel avenue: leveraging object-centric representation about principal objects in unseen domains. We demonstrate that this approach offers crucial insights for bridging the domain gap, marking a distinct departure from traditional alignment strategies.

\noindent \textbf{Action Recognition.} Action recognition, a fundamental challenge in computer vision, has seen a variety of methodological advancements. Initial approaches, like optical flow~\cite{efros2003recognizing}, laid the groundwork, followed by advancements using recurrent networks~\cite{donahue2015long,yue2015beyond}. Subsequently, the field witnessed the emergence of 3D spatio-temporal kernels~\cite{ji20133d,taylor2010convolutional,tran2015learning,varol2018long, Lin_2019_ICCV, tsn2019wang, carreira2017quo}, and two-stream networks~\cite{feichtenhofer2016convolutional,simonyan2014two, slowfast2019}, which ingeniously captured complementary signals such as motion and spatial cues. In recent times, Vision Transformers have risen to prominence, revolutionizing not only general computer vision tasks but also specifically action recognition. Pioneering works in video transformers, including ViViT~\cite{arnab2021vivit}, MViT~\cite{mvit2021}, MFormer~\cite{patrick2021keeping}, TimeSformer ~\cite{gberta_2021_ICML}, MViTv2~\cite{li2021improvedmvit}, and Video Swin~\cite{liu2021videoswin}, have demonstrated impressive performance across multiple video datasets. A notable development is ORViT~\cite{herzig2022orvit}, which posits the significant role of objects in action recognition. The proposed \Ours is inspired by ORViT that uses object-level information to adapt an action recognition model to an unseen environment. 


\Pipeline{!t}

\section{Object-based Video Domain Adaptation} \label{sec: method}

We introduce our proposed Object-based Video Domain Adaptation (\Ours), starting by describing the problem statement in~\cref{sec: problem_state} and the process to extract object-centric features in~\cref{sec: model}. In~\cref{sec: model_adapt} we describe how we adapt the object-centric features to the target environment. We give an overview of our approach in~\cref{fig: overview}.

\subsection{Preliminaries}
\label{sec: problem_state}

Given an existing action recognition model trained on a source environment $D_s$, our goal is to adapt it to an unseen target environment $D_t$, in which we aim to classify the actions of the new test video clips. Importantly, we assume access to a set of $n_t$ frames with sparse object-level annotations, $(x_f, b)\in {D_t}$, where each $x_f$ is a frame with a few associated object annotations $b$ (bounding box coordinates).



\subsection{Extracting Object-centric Feature} \label{sec: model}

The central concept of \textit{\textbf{``object-based''}} in \Ours is to adapt only the object-centric representations to the target domain, applicable across various action recognition models. \Ours comprises two key components: a class-agnostic object detector $F$ and an action recognizer $G$. 

\noindent \textbf{Obtaining the Object-centric Region.} Consider an input video segment $x = (z_1, ..., z_m) \in \mathbb{R}^{T \times H \times W \times 3}$, the  detector $F$ maps a frame $z_i\in \mathbb{R}^{H\times W\times 3}$ to a list of N class agnostic boxes (N is set to 4 in the implementation). For each frame $z_i$, each predicted box $\hat{b}_{ij} \in [0,1]^{4}$  is accompanied by a confidence score $\hat{p}_{ij}\in \mathbb{R^{d}}$.


\noindent \textbf{Enhancing the Object-centric Feature.} The region $\hat{b}$ enhances the object-centric feature within the action recognizer through $Crop$ and $Attach$ processes. For a ViT-based action recognizer, following~\cite{herzig2022orvit}, $\hat{b}$ is used to crop the original patch tokens $x_p$ to get the object tokens $x_{obj} = Crop(x_p, \hat{b})$. 
 For CNN-based models~\cite{munro2020multi, sahoo2021contrast, ta3n}, $\hat{b}$ crops the spatial feature $x_s$ extracted by their convolutional backbone. Depending on the model structure, $x_s$ varies  (\eg, for~\cite{sahoo2021contrast}, it is produced by $\mathcal{F}$, for~\cite{ta3n}, it is produced by $F^{RGB}$). This process is represented as:


\begin{align}
    x_{obj} &= \begin{cases}
        Crop(x_p, \hat{b}), & \text{if } G \text{ is a ViT-based model} \\
        Crop(x_s, \hat{b}), & \text{if } G \text{ is a CNN-based model}
    \end{cases}
\end{align}
Then, an object-centric encoder $E$ maps $x_{obj}$ to a semantically rich latent space, yielding $E(x_{obj})$. This new feature is then attached to the original $x_p$ or $x_s$ to enhance the feature for subsequent learning: $x_p := Attach(x_p, x_{obj})$ for ViT-based models and $x_s := Attach(x_s, x_{obj})$ for CNN-based models.


The action recognizer $G$ maps a video segment $x \in \mathbb{R}^{T \times H \times W \times 3}$ to an action label $G(x)\in A$. During inference, for an input video segment $x = (z_1, ..., z_m) \in \mathbb{R}^{T \times H \times W \times 3}$ in the target domain, the detector $F$ predicts frame-wise bounding boxes $[\hat{b},\hat{p}] = F(x)$. Using $\hat{b}$ and the object-centric encoder $E$, the object-centric features $E(x_{obj})$ are obtained. After attaching $E(x_{obj})$ to the original feature ($x_s$ or $x_p$), the enhanced feature propagates forward to predict the action $\hat{a} \in A$.


\subsection{Object-centric Feature Adaptation}
\label{sec: model_adapt}

\Ours \textbf{\textit{only adapts the object-centric features}} to the target domain $D_t$ by fine-tuning detector $F$ while keeping the action recognition model $G$ frozen.

\minisection{Sparse Object-level supervision}
An essential aspect of the \Ours is its reliance on object-level annotations in the target domain for fine-tuning the object detector. Despite this requirement, the annotation process is considerably more efficient compared to fully-supervised methods: (1) \textbf{\textit{\Ours requires only a sparse set of frames in total}} (we ablate this number in~\cref{sec: ablation}). This is significantly less burdensome than the standard approach, which typically employs many labeled video clips for fine-tuning, with each clip averaging around 179 frames in datasets like \textit{Epic-Kitchens} (refer to~\cref{sec: sup} for details). (2) The annotation effort is further streamlined by \textbf{\textit{focusing exclusively on hands and objects interacting with hands}}, thus reducing the complexity and volume of required bounding box annotations (we do not require the object labels). (3) The process of annotating these frames can be greatly \textbf{\textit{facilitated by foundational segmentation models such as SAM}}~\cite{kirillov2023segment}, that makes a bounding box annotation equal to a mouse click. While models like SAM are instrumental in frame-level annotation, they are not equipped for temporal labeling, hence they are unable to annotate full video clips. 

\noindent \textbf{Adaptation loss.} Intuitively, this adaptation method deals with the uncertainty related to encountering previously unseen objects in the target environment, as well as other factors like the background and lightning conditions.
We adapt the object-centric feature by fine-tuning $F$ using the sparsely labeled frames $(x_f, b)\in {D_t}$. 
We minimize the following objective:

\vspace{-5mm}
\begin{equation}
\mathcal{L} := \sum_{i \in T'} \sum_{j=1}^4 \mathcal{L}_{CE}(\sigma(\hat{p}_{ij}), p_{ij}) + p_{ij} (L_{1}(\hat{\bb}_{ij}, \bb_{ij}))
\label{eq:loss_objects}
\end{equation}
where $\sigma$ is the sigmoid function, $\mathcal{L}_{CE}$ is the cross-entropy loss, $p_{ij}\in \{0,1\}$ is a binary indicator that receives $1$ if the ${j^{th}}$ object predicted in the ${i^{th}}$ frame overlaps with the ground truth object based on a specific IoU threshold.


During inference, target video clips are input into the adapted object detection model to predict the object-centric regions, which are then utilized for enhancing the object-centric features. This enhancement facilitates the calculation of the predicted actions, as illustrated in~\cref{fig: overview}.

\def\TabEpic#1{
\begin{table*}[#1]
    \tablestyle{11.pt}{1.0}
    \small
    \begin{center}
    \begin{tabular}{l|cccccc|l}
    Method                             & D2→ D1 & D3→ D1 & D1→ D2 & D3→ D2 & D1→ D3 & D2→ D3 & Mean \\ \shline
    Source only & 33.0        &33.9        & 34.4   &42.4        & 34.4   & 41.7   & 36.6     \\
    Fully-supervised & 54.3 & 56.2 & 57.9 & 61.0 & 52.5 & 58.2 & 56.7 \\ \hline
    \rowcolor{gray!10}
    \Ours  & 41.0   & 42.2   & 43.8   & 47.9   & 39.3   & 44.5   & 43.1 \\
    \rowcolor{gray!10} \textbf{$\Delta$}                 & \textcolor{ForestGreen}{\textbf{+8.1}}       & \textcolor{ForestGreen}{\textbf{+8.4}}       & \textcolor{ForestGreen}{\textbf{+9.4}}       & \textcolor{ForestGreen}{\textbf{+5.5}}       &  \textcolor{ForestGreen}{\textbf{+4.9}}      &  \textcolor{ForestGreen}{\textbf{+2.8}}      &  \textcolor{ForestGreen}{\textbf{+6.5}}    \\ 
    
    \end{tabular} \vspace{-1.2em}
    \end{center}
    \caption{\textbf{Adaptation within dataset.} We report the top-1 accuracy of 8 action classes on different kitchen pairs \textbf{within \textit{Epic-Kitchens}}. The $\Delta$ denotes the absolute boost between the source only model and adaptation model. In the adaptation process, we use 100 frames in the target kitchens to fine-tune the detector, and use DINO encoder for object region.}
    \label{tab:ek_main}
\end{table*}
}

\def\TabEGTEA#1{
\begin{table*}[#1]
    \tablestyle{4.pt}{1.0}
    \small
    \begin{center}
    \begin{tabular}{l|ccccccc|l}
    Method              & GreekSalad & Pizza & Cheeseburger & BaconAndEgg & TurkeySandwich & ContinentalBreak & PastaSalad & Mean \\ \shline
    Source only & 41.0       & 23.0  & 35.3         & 35.8        & 36.7           & 35.3             & 42.6       & 35.7 \\
    Fully-supervised & 57.6 & 39.3 & 49.7 & 56.2 & 49.7 & 50.4 & 52.3 & 50.6\\ \hline
    \rowcolor{gray!10}
    \Ours  & 45.4       & 26.0  & 37.2         & 39.2        & 39.3           & 38.7             & 45.2       & 38.7 \\
    \rowcolor{gray!10} \textbf{$\Delta$}       & \textcolor{ForestGreen}{\textbf{+4.4}}        & \textcolor{ForestGreen}{\textbf{+3.1}}   & \textcolor{ForestGreen}{\textbf{+1.9 }}         & \textcolor{ForestGreen}{\textbf{+3.3}}         & \textcolor{ForestGreen}{\textbf{+2.6}}           & \textcolor{ForestGreen}{\textbf{+3.4}}             & \textcolor{ForestGreen}{\textbf{+2.6}}        & \textcolor{ForestGreen}{\textbf{+3.1}}  \\
    
    \end{tabular} \vspace{-1.2em}
    \end{center}
    \caption{\textbf{Adaptation across datasets.} We report the top-1 accuracy of multiple action classes on different recipes in \textbf{\textit{EGTEA} adapted from \textit{Epic-Kitchens}}. The $\Delta$ denotes the absolute boost between the source only model and adaptation model. In the adaptation process, we use 100 frames in the target recipe to fine-tune the detector, and use DINO encoder for object region.}
    \label{tab: egtea_main}
    
\end{table*}
}

\def\TabComparison#1{
\begin{table*}[#1]
    \tablestyle{12.pt}{1.0}
    \small
    \begin{center}
    \begin{tabular}{l|cccccc|l}
    Method                             & D2→ D1 & D3→ D1 & D1→ D2 & D3→ D2 & D1→ D3 & D2→ D3 & Mean \\ \shline
    
    & \multicolumn{7}{c}{\textbf{\textit{ViT-based methods}}}                                         \\
    
    Source only & 33.0        &33.9        & 34.4   &42.4        & 34.4   & 41.7   & 36.6     \\
    Fully-supervised & 54.3 & 56.2 & 57.9 & 61.0 & 52.5 & 58.2 & 56.7 \\ \hline
    \rowcolor{gray!10}
    \Ours  & 41.0   & 42.2   & 43.8   & 47.9   & 39.3   & 44.5   & 43.1 \\
    \rowcolor{gray!10} \textbf{$\Delta$}                 & \textcolor{ForestGreen}{\textbf{+8.0}}       & \textcolor{ForestGreen}{\textbf{+8.3}}       & \textcolor{ForestGreen}{\textbf{+9.4}}       & \textcolor{ForestGreen}{\textbf{+5.5}}       &  \textcolor{ForestGreen}{\textbf{+4.9}}      &  \textcolor{ForestGreen}{\textbf{+2.8}}      &  \textcolor{ForestGreen}{\textbf{+6.5}}    \\ \shline

                 & \multicolumn{7}{c} {\textbf{\textit{Methods based on adversarial learning}}}                                   \\
    Source only  & 35.4   & 34.6   & 32.8   & 35.8   & 34.1   & \multicolumn{1}{c|}{39.1}   & 35.3 \\
    DANN~\cite{dann}         & 38.3   & 38.8   & 37.7   & 42.1   & 36.6   & \multicolumn{1}{c|}{41.9}   & 39.2 \\
    ADDA~\cite{adda}         & 36.3   & 36.1   & 35.4   & 41.4   & 34.9   & \multicolumn{1}{c|}{40.8}   & 37.5 \\
    TA$^3$N\cite{ta3n}         & 40.9   & 39.9   & 34.2   & 44.2   & 37.4   & \multicolumn{1}{c|}{42.8}   & 39.9 \\ \hline
    \rowcolor{gray!10} TA$^3$N-\Ours   & 41.4   & 42.3   & 38.9   & 45.4   & 36.8   & \multicolumn{1}{c|}{44.8}   & 41.6 \\
    \rowcolor{gray!10} $\Delta$  & \textcolor{ForestGreen}{\textbf{+0.5}}   & \textcolor{ForestGreen}{\textbf{+2.4}}   & \textcolor{ForestGreen}{\textbf{+4.7}}   & \textcolor{ForestGreen}{\textbf{+1.2}}   & \textcolor{BrickRed}{\textbf{-0.6}}   & \multicolumn{1}{c|}{\textcolor{ForestGreen}{\textbf{+2.0}}}   & \textcolor{ForestGreen}{\textbf{+1.7}} \\ \shline
                 
                 & \multicolumn{7}{c}{\textbf{\textit{Methods based on contrastive learning}}}                                   \\
    Source only  & 35.4   & 34.6   & 32.8   & 35.8   & 34.1   & \multicolumn{1}{c|}{39.1}   & 35.3 \\
    CoMix~\cite{sahoo2021contrast}        & 38.6   & 42.3   & 42.9   & 49.2   & 40.9   & \multicolumn{1}{c|}{45.2}   & 43.2 \\ \hline
    \rowcolor{gray!10} CoMix-\Ours  & 39.3   & 41.4   & 44.7   & 50.2   & 41.2   & \multicolumn{1}{c|}{46.7}   & 43.9 \\
    \rowcolor{gray!10} $\Delta$             & \textcolor{ForestGreen}{\textbf{+0.7}}   & \textcolor{BrickRed}{\textbf{-0.9}}   & \textcolor{ForestGreen}{\textbf{+1.8}}   & \textcolor{ForestGreen}{\textbf{+1.0}}   & \textcolor{ForestGreen}{\textbf{+0.3}}   & \multicolumn{1}{c|}{\textcolor{ForestGreen}{\textbf{+1.5}}}   & \textcolor{ForestGreen}{\textbf{+0.7}} \\ \shline

    & \multicolumn{7}{c}{\textbf{\textit{Multi-modal methods}}}                                         \\
    Source only  & 42.5   & 44.3   & 42.0   & 56.3   & 41.2   & \multicolumn{1}{c|}{46.5}   & 45.5 \\
    MMSADA~\cite{munro2020multi}       & 48.2   & 50.9   & 49.5   & 56.1   & 44.1   & \multicolumn{1}{c|}{52.7}   & 50.3 \\ \hline
    \rowcolor{gray!10} MMSADA-\Ours & 51.5   & 55.9   & 55.2   & 61.9   & 50.3   & \multicolumn{1}{c|}{56.8}   & 55.3 \\
    \rowcolor{gray!10} $\Delta$             & \textcolor{ForestGreen}{\textbf{+3.3}}   & \textcolor{ForestGreen}{\textbf{+5.0}}   & \textcolor{ForestGreen}{\textbf{+5.7}}   & \textcolor{ForestGreen}{\textbf{+5.8}}   & \textcolor{ForestGreen}{\textbf{+6.2}}   & \textcolor{ForestGreen}{\textbf{+4.1}}                        & \textcolor{ForestGreen}{\textbf{+5.0}} \\
                 
    \end{tabular} \vspace{-1.2em}
    \end{center}
    \caption{\textbf{Adaptation within dataset.} We report the top-1 accuracy of 8 action classes on different kitchen pairs \textbf{within \textit{Epic-Kitchens}}. We show the performance of \Ours when applied on top of different action recognition models. The $\Delta$ denotes the absolute boost between the source only model and adaptation model.  In the adaptation process, we use 100 frames in the target kitchens to fine-tune the detector, and use DINO encoder for object region.}
    \label{tab: comp}
\end{table*}
}

\def\FigVis#1{
\begin{figure*}[#1]
\begin{center}
\includegraphics[width=\linewidth]{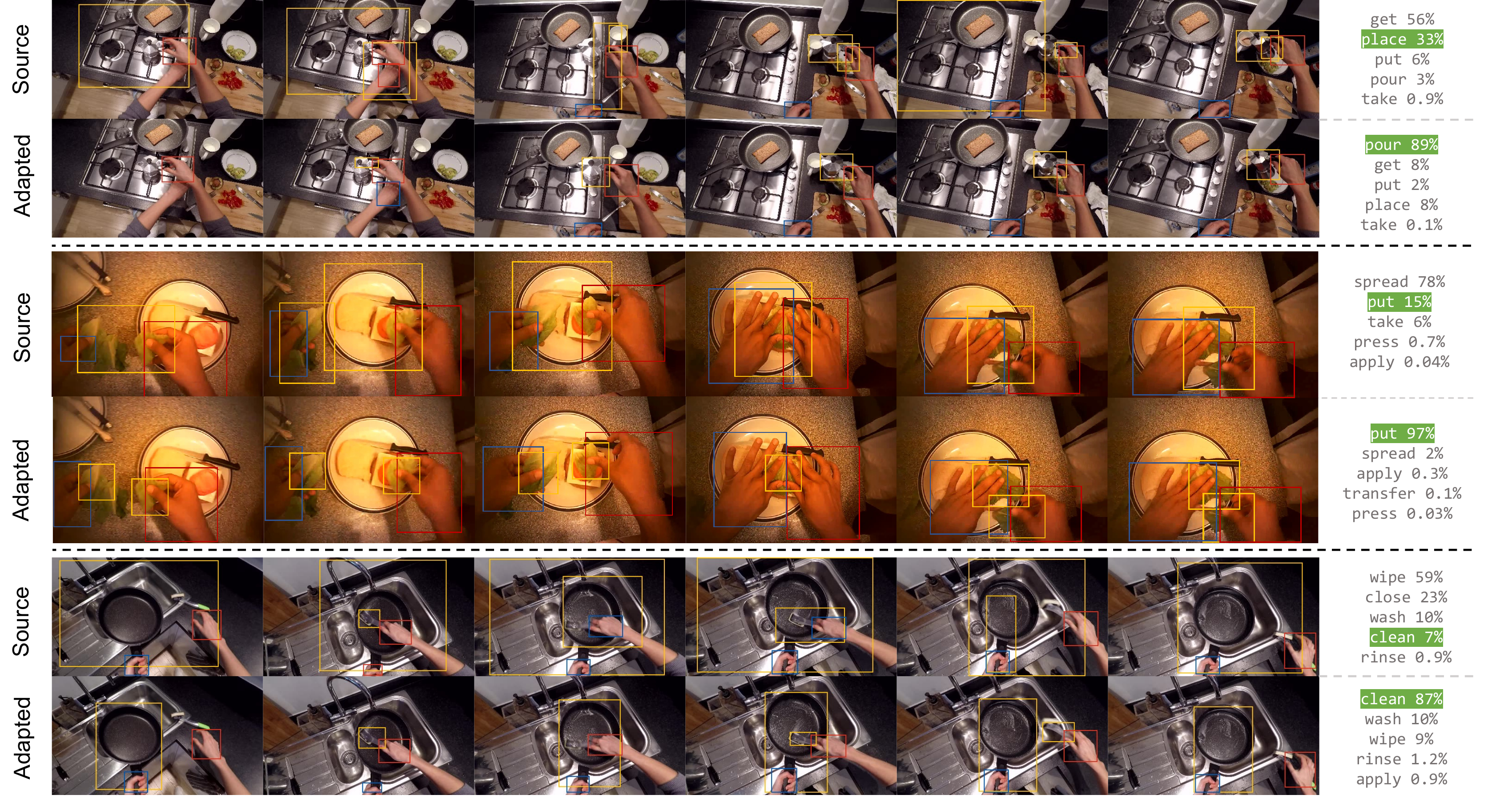}
\end{center}
\vspace{-0.5cm}
\caption{\textbf{Visualization of the adaptation using only 100 labeled frames.} We show three examples of adaptations across datasets. The source action recognition is trained on \textit{Epic-Kitchens} and the three test example are from target kitchens in \textit{Epic-Kitchens} and recipes in \textit{EGTEA}. For each
example, we include predicted objects from source detector (top) and adapted detector (down). Fed with different objects, we show the predicted top-5 actions with their confidence score, and the corresponding ground truth (marked as green). }
\label{fig: qualitative_egtea}
\end{figure*}
}

\def\FigFt#1{
\begin{figure*}[#1]
\begin{center}
\includegraphics[width=\linewidth]{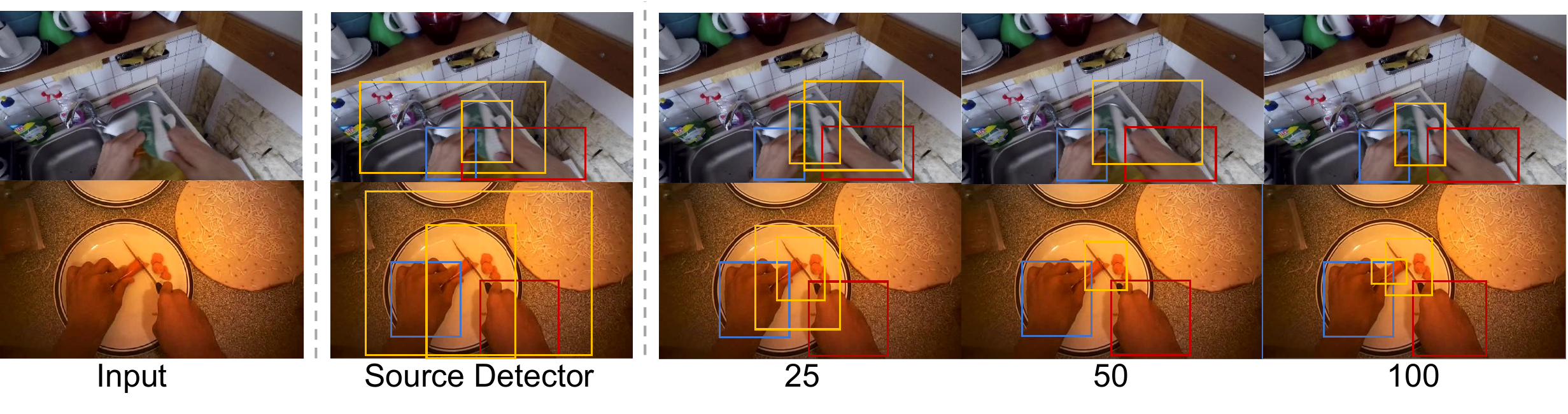}
\end{center}
\vspace{-0.5cm}
\caption{Visualizations for fine-tuning of the object detector with \textbf{different number of frames}. We visualize the bounding boxes predicted by the source detector and the adapted detector fine-tuned with different number of frames in the target domain. We adapt kitchens across \textit{Epic-Kitchens} (the top), and kitchens in \textit{Epic-Kitchens} to a recipe in \textit{EGTEA} (down).}
\label{fig: qualitative_boxes}
\end{figure*}
}

\def\FigFtActionR#1{
\begin{figure}[#1]
\begin{center}
\includegraphics[width=\linewidth]{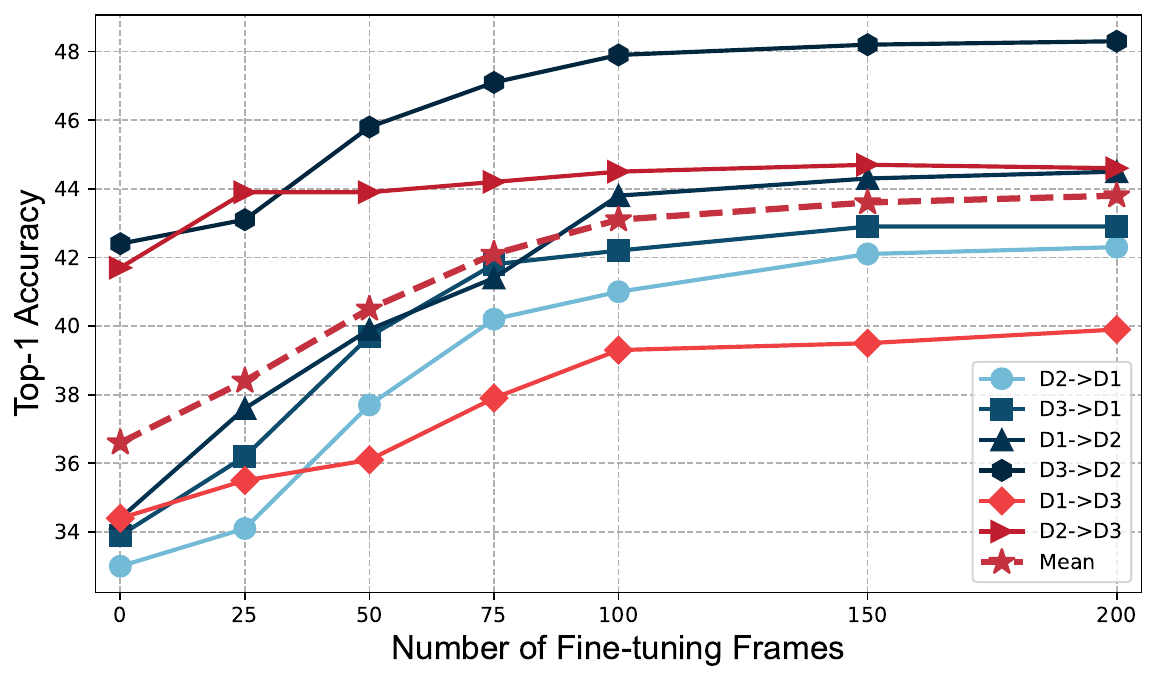}
\end{center}
\vspace{-0.5cm}
\caption{The performance of \Ours using \textbf{different number of frames} to fine-tune the object-detector. We report the results of the adaptation within \textit{Epic-Kitchens} dataset, and ViT-based action recognition model with DINO as encoder.}
\label{fig: ftnum_acc}
\end{figure}
}

\def\TabEncoder#1{
\begin{table*}[#1]
    \tablestyle{7.pt}{1.0}
    \small
    \begin{center}
    \begin{tabular}{l|ccccccc|l}
    Method        & Encoder              & D2→ D1        & D3→ D1        & D1→ D2        & D3→ D2        & D1→ D3        & D2→ D3        & Mean          \\ \shline
    
    Source only                 & & 33.0          & 33.9          & 34.4          & 42.4          & 34.4          & 41.7          & 36.6          \\ 
    \hline
    
    \multirow{4}{*}{\Ours} &   \usym{2717}                 & 35.5          & 37.7          & 36.2          & 45.0          & 35.3          & 42.4          & 38.7        
    
    \\
                              & SwAV \cite{swav}                & 32.4          & 35.6          & 36.7          & 45.9          & 38.9          & 42.8          & 38.7 \textcolor{ForestGreen}{\footnotesize (+0.0)}          \\
                                & CLIP  \cite{clip}               & 38.7          & 39.2          & 40.2          & 46.2          & 39.1          & 41.0          & 40.8 \textcolor{ForestGreen}{\footnotesize (+2.1)}          \\
                                \rowcolor{gray!10} & DINO \cite{dino}                & \textbf{41.0} & \textbf{42.2} & \textbf{43.8} & \textbf{47.9} & \textbf{39.3} & \textbf{44.5} & \textbf{43.1 }\textcolor{ForestGreen}{\footnotesize (+3.4)} \\ 
    \end{tabular}
    \caption{The performance of \Ours with \textbf{different object encoders}. We report the results of the adaptation within \textit{Epic-Kitchens} dataset with different object encoder. We use ViT-based action recognition model and 100 frames in fine-tuning.}
    \label{tab: obj_encoder}
    \end{center}
\end{table*}
}

\def\TabFt#1{
\begin{table*}[#1]
    \tablestyle{7.pt}{1.0}
    \small
    \begin{center}
    \begin{tabular}{l|ccccccc|l}
    & \begin{tabular}[c]{@{}c@{}}\# Fine-tuning\\ frames\end{tabular} & D2→ D1 & D3→ D1 & D1→ D2 & D3→ D2        & D1→ D3        & D2→ D3        & Mean          \\ \shline
    Source only                 & 0                                                                & 33.0          & 33.9          & 34.4          & 42.4          & 34.4          & 41.7          & 36.6          \\ \hline
    \multirow{4}{*}{Adaptation} & 25                                                              & 34.1          & 36.2          & 37.6          & 43.1          & 35.5          & 43.9          & 38.4 \textcolor{ForestGreen}{\footnotesize (+1.8)}          \\
                                & 50                                                              & 37.7          & 39.7          & 39.9          & 45.8          & 36.1          & 43.9          & 40.5 \textcolor{ForestGreen}{\footnotesize (+3.9)}          \\
                                & 75                                                              & 40.2          & 41.8          & 41.4          & 47.1          & 37.9          & 44.2          & 42.1 \textcolor{ForestGreen}{\footnotesize (+5.5)}          \\ \rowcolor{gray!10}
                                & 100 &\textbf{41.0} & \textbf{42.2} & \textbf{43.8} & \textbf{47.9} & \textbf{39.3} & \textbf{44.5} & \textbf{43.1} \textcolor{ForestGreen}{\footnotesize (+6.5)} \\ 
    \end{tabular} \vspace{-1.2em}
    \end{center}
    \caption{The performance of \Ours with \textbf{different number of frames} in the fine-tuning of the object-detector. We report the results of the adaptation within \textit{Epic-Kitchens} dataset with different object encoder. \amir{Change to a line plot. We need to emphasize that after X small number of labeled frames, more labeled data leads to marginal improvements}} 
    \label{tab: ablate_number}
\end{table*}
}

\TabComparison{!t}
\TabEGTEA{!t}
\FigVis{!ht}

\section{Experiments and Results}
\subsection{Experimental setup}
\label{exp:setup} 
\minisection{Experiment Design}
We design experiments in  two scenarios: \textbf{\textit{(1) Adaptation to an unseen environment within a dataset}}. We follow the existing benchmark established in \cite{munro2020multi, sahoo2021contrast} on the \textit{Epic-Kitchens} dataset~\cite{Damen2018EPICKITCHENS,damen2020epic}, which takes the three most frequent kitchens as separate domains and performs adaptation across them. (2) We construct a more challenging scenario of \textbf{\textit{adapting not only to an unseen kitchen but also to a new dataset.}} We use \textit{Epic-Kitchens} as source data and \textit{EGTEA Gaze+ videos}~\cite{li2018eye} as targets. 


\minisection{Implementation details} Our method is implemented with PyTorch. We train the models on $8$ NVIDIA Quadro RTX 6000 GPUs. We implement our idea for ViT-based action recognition models and the state-of-the-art CNN-based models commonly used in self-/unsupervised video-based domain adaptation~\cite{sahoo2021contrast,ta3n,munro2020multi}.

\minisection{Action Recognizer $G$} For the ViT-based action recognizer, following the standard training schedule in~\cite{herzig2022orvit}, we train the source model for 40 epochs with a learning rate of $10^{-4}$. For the CNN-based action recognizer, we follow the design of~\cite{sahoo2021contrast,ta3n,munro2020multi} to apply our method on top of them. \textbf{\textit{The action recognizer is frozen in the adaptation period.}}

\minisection{Object Detector $F$}In our experiment, we use~\cite{Shan20} as an class-agnostic object detector (\ie the category and hand-object interaction prediction is not used) to localize the object-cnetric region. During training, we use frames with bounding boxes in the source domain to train a detector. During adaptation, we first use 100 annotated frames (we ablate this number in~\cref{sec: ablation}) from the target domain to fine-tune the detector. In the inference time, the object detector is replaced with the fine-tuned one that is better aligned with the target domain.

\minisection{Object-centric Encoder $E$} We leverage the strong feature extraction abilities of the recent self-supervised pre-trained models SwAV~\cite{swav}, DINO~\cite{dino} and CLIP~\cite{clip} to encode the object regions to strengthen their representation. In ~\cref{sec: 4.2} and~\ref{sec: 4.3}, we show the results with using DINO as an encoder, which achieves the best performance for ViT-based models. In~\cref{sec: comp_adapted}, we use SwAV for CNN-based models. We compare and ablate the encoders in~\cref{sec: ablation}.

\minisection{Datasets} We conduct experiments on \textit{Epic-Kitchens}~\cite{damen2020epic} and \textit{EGTEA Gaze+}~\cite{li2018eye}. \textit{Epic-Kitchens} is a large-scale dataset for first-person (egocentric) vision, including different kitchens. \textit{EGTEA Gaze+ (EGTEA)} is another large dataset for egocentric vision actions that spans 7 recipes. For training the object detector we use the annotations from the \textit{100K Frame}~\cite{shan2020understanding} (\textit{100K}), which is built upon \textit{100 Days Of Hands} (\textit{100DOH}).


\minisection{Baselines} For baselines, we use: (1) a source-only model (a lower bound) trained solely in the source domain; (2) a fully-supervised model (an upper bound) that uses a comparable\footnote{Comparable number refers to the same number of frames we use to fine-tune our detector during adaptation.} number of video clips in the target domain to do end-to-end fine-tuning of the source-only model during adaptation; (3) popular unsupervised domain adaptation (UDA) models based on adversarial learning~\cite{dann, adda, ta3n}; (4) the state-of-the-art video domain adaptation model using contrastive ~\cite{sahoo2021contrast} and multi-modal~\cite{munro2020multi} learning architectures.

\minisection{Evaluation Metrics} For the first scenario (adaptation across kitchens within \textit{Epic-Kitchens}), following~\cite{munro2020multi, sahoo2021contrast}, we report the top-1 accuracy of 8 most common action classes\footnote{Consistent with prior work~\cite{munro2020multi, sahoo2021contrast}, by actions here we just mean the verbs, as annotated in the EK dataset.} from three domains named D1, D2 and D3, corresponding to P08, P01 and P22 kitchens in the \textit{Epic-Kitchens} dataset. For the second scenario (adaptation from \textit{Epic-Kitchens} to \textit{EGTEA}), we report the top-1 accuracy on 19 overlapping action classes of these two datasets (see~\cref{sec: domain_gap} for details).

\subsection{Adaptation within a Dataset} \label{sec: 4.2}
Adhering to the established benchmark using the \textit{Epic-Kitchens} dataset~\cite{munro2020multi, sahoo2021contrast}, our study aims to assess the adaptability of an action recognition model to a novel kitchen environment, maintaining consistency in camera devices, recording setups, and annotation strategies. In this context, varying elements such as lighting, background, object types, and participant behaviors contribute to a decline in model performance.

\noindent \textbf{Results.} As detailed in~\cref{tab: comp} (\emph{ViT-based methods}), results reveal that \Ours significantly enhances performance, achieving an average increase of \textbf{+6.5} in top-1 accuracy across six different kitchen pairings. We juxtapose our results against a fully-supervised counterpart, deemed as the performance upper-bound. Here, instead of utilizing all available clips in the target domain for fine-tuning, we employ only those clips that correspond to the sparsely annotated frames in \Ours. This approach, focusing on a ``frame \vs clip'' comparison, lays the groundwork for exploring the potential of weak- or unsupervised learning methods: specifically, \textbf{\textit{investigating the feasibility of substituting costly clip-level supervision with cheaper object-level supervision}}. Our findings indicate that a significant portion of the performance degradation when transitioning from the source to the target domain is due to the model's inability to accurately capture object information in an unfamiliar environment. Addressing this, we demonstrate that employing an enhanced object detector as part of our framework effectively narrows this domain gap.

In addition, we present some qualitative results in~\cref{fig: qualitative_egtea}, that showcases three target video clips alongside their object-level predictions from both the source and adapted detectors, as well as the corresponding action predictions. A notable observation is that enhanced object region detection leads to more precise action predictions. Take, for instance, the ``pour the coffee'' scenario (illustrated in the first and second columns). The source detector incorrectly assumes that \textit{the left hand is holding something}, consistently producing false positives around the oven area. In the contrast, the adapted detector consistently avoids such errors. Furthermore, with the adapted detector, the recognition of the object interacted with the right hand near the coffee container is remarkably stable both temporally and spatially. As a result, the action recognizer can accurately identify the action as ``pouring'' rather than resorting to more generic actions like ``get'' or ``place''.

\subsection{Adaptation across Datasets} \label{sec: 4.3}
In this particular scenario, we perform domain adaptation from the three kitchens in the \textit{Epic-Kitchens} dataset to the seven distinct recipes from the \textit{EGTEA} dataset. This setup presents a notably challenging environment for domain adaptation, primarily due to the divergences in recording instructions, hardware used, and annotation protocols between these two datasets. Our objective is to rigorously evaluate the performance of our method when confronted with an expanded domain gap, testing its robustness and adaptability in more complex and varied environments.

\noindent \textbf{Results.} The results for this scenario, as depicted in~\cref{tab: egtea_main}, demonstrate the effectiveness of our method in bridging a substantial domain gap. Notably, our approach achieves an average performance boost of \textbf{+3.1}, with a remarkable improvement of \textbf{+4.4} in recognizing actions related to preparing a ``Greek Salad''. The qualitative outcomes are illustrated as the second example ``put lettuce'' in~\cref{fig: qualitative_egtea}. 


\subsection{Comparison and Adaptation to the SOTAs} \label{sec: comp_adapted}
Next, we combine our ``object-based'' idea with the state-of-the-art video domain adaptation approaches.

\noindent \textbf{Methods based on adversarial learning.} \label{sec: ala}
Among various strategies, such as DANN~\cite{dann}, ADDA~\cite{adda}, SAVA~\cite{choi2020shuffle}, and TA$^3$N~\cite{ta3n}, we specifically apply our approach on top of TA$^3$N (see details for CNN-based models in~\cref{sec: method}), given its superior performance on the \textit{Epic-Kitchens} setting. 


We present the top-1 accuracy for 8 actions using both the original adversarial learning-based model and the combined TA$^3$N-\Ours model in~\cref{tab: comp} (\emph{Methods based on adversarial learning}). The adaptation with the ``object-based'' approach results in a notable performance enhancement, yielding an average increase of \textbf{+1.7} in accuracy compared to the original TA$^3$N. These outcomes align with those from our ViT-based model, underscoring the pivotal role of object representation in action recognition. The improved representation of objects in an unseen domain effectively aids in reducing the domain gap, thereby enhancing model performance in diverse environments.

\noindent \textbf{Methods based on contrastive learning.}
In the realm of models based on contrastive learning, we extend our ``object-based'' approach to CoMix~\cite{sahoo2021contrast}, a leading method in the \textit{Epic-Kitchens} setting. 
The comparative results, as shown in~\cref{tab: comp} (\emph{Methods based on contrastive learning}), indicate an average performance boost of \textbf{+0.7} for the combined CoMix-\Ours, slightly lower than the enhancement seen in the adapted method based on adversarial learning.

\noindent \textbf{Multi-modal methods.} Focusing on multi-modal methods, MMSADA~\cite{munro2020multi} stands out for its high performance in the \textit{Epic-Kitchens} benchmark, largely attributable to their inclusion of a flow branch. We modify the RGB branch of MMSADA, employing our established pipeline to crop, encode, and attach object representations to the features extracted from their convolutional backbone. As detailed in~\cref{tab: comp} (\emph{Multi-modal methods}), the combined MMSADA-\Ours surpasses the original MMSADA model, achieving an average improvement of \textbf{+5.0}, marking a substantial leap from the Source-only model.

In summary, the implementation of our ``object-based'' idea not only empowers ViT-based models to effectively bridge domain gap but also, when combined with the state-of-the-art video domain adaptation models built on CNN architectures, yields further performance enhancements. This versatility underscores its broad applicability and efficacy.

\TabEncoder{!t}

\FigFtActionR{!t}

\subsection{Ablation Studies} \label{sec: ablation}
In this section, we conduct ablation studies to scrutinize two vital components integral to the \Ours framework: (1) the efficacy of the object encoder; (2) the influence of varying the number of frames used in the fine-tuning of the object detector. These experimental analyses are performed in the context of our first scenario: adaptation across kitchens within \textit{Epic-Kitchens} dataset. The objective of these studies is to isolate and understand the individual contributions of these components to the overall performance of \Ours. 

\FigFt{!ht}

\noindent \textbf{Object-centric Encoder.} \label{sec: obj_encoder}
To effectively leverage the rich object information available to us, our approach incorporates the use of an object encoder. The primary function of this encoder is to transform low-level features into a latent space enriched with semantically informative content. We employed several cutting-edge self-supervised pre-trained models in our experiments. These include: (1) SwAV~\cite{swav}, utilizing a ResNet-50 backbone for feature extraction; (2) DINO~\cite{dino}, which employs the last 6 blocks of a ViT-small architecture with a 16 $\times$ 16 patch size; (3) CLIP~\cite{clip}, using the last 4 blocks of a visual ViT-small, also with a 16 $\times$ 16 patch size. The comparative results of these models, in the context of our framework, are presented in~\cref{tab: obj_encoder}.

Among the evaluated models, DINO emerged as the most effective. In contrast, SwAV exhibited minimal improvement in our framework. This disparity in performance can be attributed to the architectural differences: SwAV, with its CNN-based backbone, is less adept at extracting nuanced information compared to the large ViT-based structure of our action recognition model. It is important to note that for maintaining consistency within the model architecture, SwAV was exclusively employed as the object-centric encoder in all CNN-based domain adaptation methods discussed in~\cref{sec: comp_adapted}.



\noindent \textbf{Number of Fine-tuning Frames.}
The number of frames for fine-tuning the object detector in the target domain is a critical factor influencing the adaptation performance. To investigate this, we conduct experiments with varying frame quantities (25, 50, 75, 100, 150 and 200) for object detector fine-tuning (all the shown results in previous sections use 100 frames). The results, as illustrated in~\cref{fig: ftnum_acc}, reveal a clear positive correlation between the number of fine-tuning frames and the accuracy of the action prediction, and the benefits begin to plateau with higher frame counts. Specifically, the performance increment from 75 to 100 frames shows signs of saturation, suggesting that 100 frames strike a reasonable balance between performance gains and resource expenditure.


\cref{fig: qualitative_boxes} showcases qualitative results of fine-tuning the object detector. In the first example, we selected an image from kitchen D2 in the \textit{Epic-Kitchens} dataset that was not part of our fine-tuning set. The comparison of the object detection predictions from the source detector and adapted detectors fine-tuned with different numbers of annotated frames is displayed. We could see increasing number of fine-tuning frames enables the object detector to better focus on nuanced feature.
For example, in the ``cut carrots'' example shown in the second row, the source detector struggles with accurate hand-interacted object localization. However, post fine-tuning in the target environment, the detector's capability to precisely localize the interacted object markedly improves. This improvement is evident in the enhanced distinction of the object (``carrot'', ``knife'') from its surroundings (``counter'', ``plate''), highlighting the value of fine-tuned object detection in providing more granular and accurate contextual information.

\section{Limitations}


\Ours strives to mitigate the domain gap by adapting the object-centric features without relying on extensively annotated video data. Although \Ours uses a class-agnostic object detector, which does not require the source and target domains to share the same objects, similar to other unsupervised methods~\cite{munro2020multi, ta3n, sahoo2021contrast, adda, dann, choi2020shuffle}, \Ours primarily targets scenarios with overlapping actions between source and target domains. This limitation stems from \Ours not utilizing a new set of video data in the target domain, which may include novel actions, for fine-tuning the action recognition model. However, action recognition models in practical applications are typically trained on broad source domains, suggesting that adaptation is often from a broad to a specific domain, with the former encompassing the latter. Additionally, generating synthetic data that closely mirrors new actions not present in the training dataset can be a viable strategy to address this limitation.

\section{Conclusion}

In this work, we tackled the practical challenge of adapting an existing action recognition model to a new, unseen environment. Central to our approach is the premise that objects significantly contribute to bridging the domain gap. Building on this insight, we introduced \Ours, a straightforward yet effective paradigm that enhances domain adaptation by enriching the action recognition model with improved object information in the target domain. Our investigation spans two demanding scenarios: ``within'' and ``across'' dataset adaptations, utilizing large-scale egocentric cooking benchmarks. The results demonstrate that \Ours consistently enhances the performance of ViT-based models. Moreover, when integrated with the state-of-the-art CNN-based video domain adaptation models, \Ours further boosts performance. The implications of this study extend beyond its immediate findings, opening promising avenues for future research in domain adaptation, few-shot learning, and unsupervised learning within the context of action recognition. We believe our work lays a foundation that will inspire and guide researchers exploring these interconnected fields.

{
    \small
    \bibliographystyle{ieeenat_fullname}
    \bibliography{main}
}


\def\LISA#1{
\begin{table*}[#1]
    \tablestyle{14.pt}{1.0}
    \small
    \begin{center}
    \begin{tabular}{l|cccccc|l}
     & D2→ D1 & D3→ D1 & D1→ D2 & D3→ D2        & D1→ D3        & D2→ D3        & Mean          \\ \shline
    Source only                                                                                 & 33.0          & 33.9          & 34.4          & 42.4          & 34.4          & 41.7          & 36.6          \\ \hline
    LISA                                                               & 35.2          & 37.1          & 38.6          & 43.7          & 36.5          & 44.2          & 39.2           \\ LISA+ & 38.2          & 40.0          & 40.3          & 46.2          & 36.8          & 44.4          & 41.0          \\
    SAM + Point & 40.8          & 42.2          & 43.6          & 48.0          & 39.0          & 44.5          & 43.0          \\
    
                        \hline \rowcolor{gray!10}       Manual  &\textbf{41.0} & \textbf{42.2} & \textbf{43.8} & \textbf{47.9} & \textbf{39.3} & \textbf{44.5} & \textbf{43.1}  \\ 
    \end{tabular} \vspace{-1.2em}
    \end{center}
    \caption{The performance of \Ours using \textbf{automatically labeling}.We assess the performance of \Ours using automatically generated segmentation that are subsequently converted into bounding boxes, which are then utilized to fine-tune the class-agnostic object detector during the domain adaptation phase. ``LISA'' denotes the use of the LISA model to segment 100 frames for evaluation purposes. ``LISA+'' indicates an post-segmentation step where noisy labels are filtered out before evaluation. ``SAM + Point'' refers to the use of a point-based SAM model to generate the labels.}

    \label{tab: automatic_label}
\end{table*}
}

\def\SuppSAM#1{
\begin{figure*}[#1]
\begin{center}
\includegraphics[width=\linewidth]{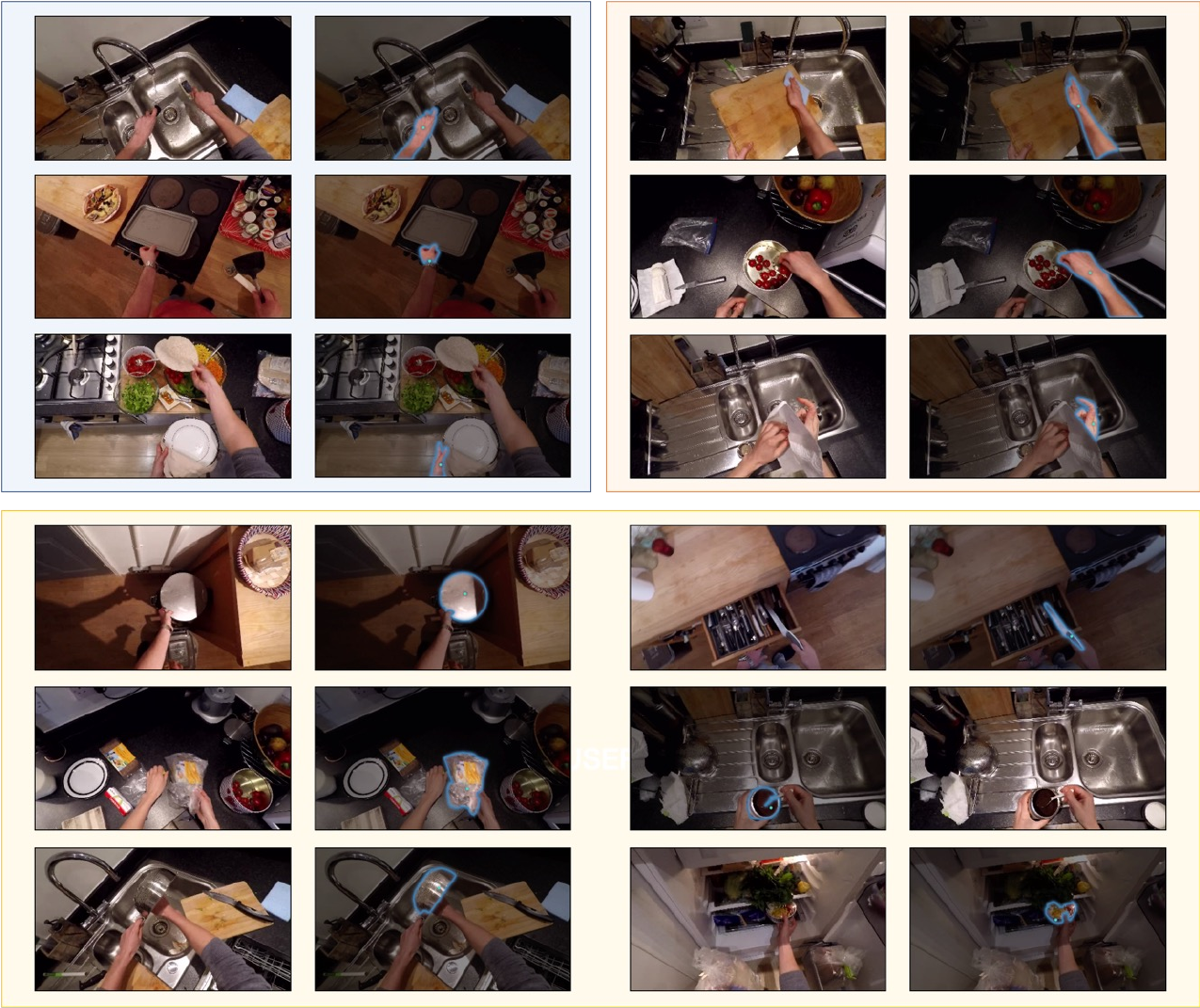}
\end{center}
\vspace{-0.5cm}
\caption{\textbf{The segmentation results by prompting SAM~\cite{kirillov2023segment} with a mouse click.} We prompt SAM with mouse clicks to get the segmentations of  ``objects interacted with hands'' (yellow boxes), ``left hand'' (blue boxes) and ``right hand'' (red boxes). During the domain adaptation phase of \Ours, these segmentations could serve as supervision to fine-tune the class-agnostic object detector.} 
\label{fig: supp_sam}
\end{figure*}
}

\def\SuppLISA#1{
\begin{figure*}[#1]
\begin{center}
\includegraphics[width=\linewidth]{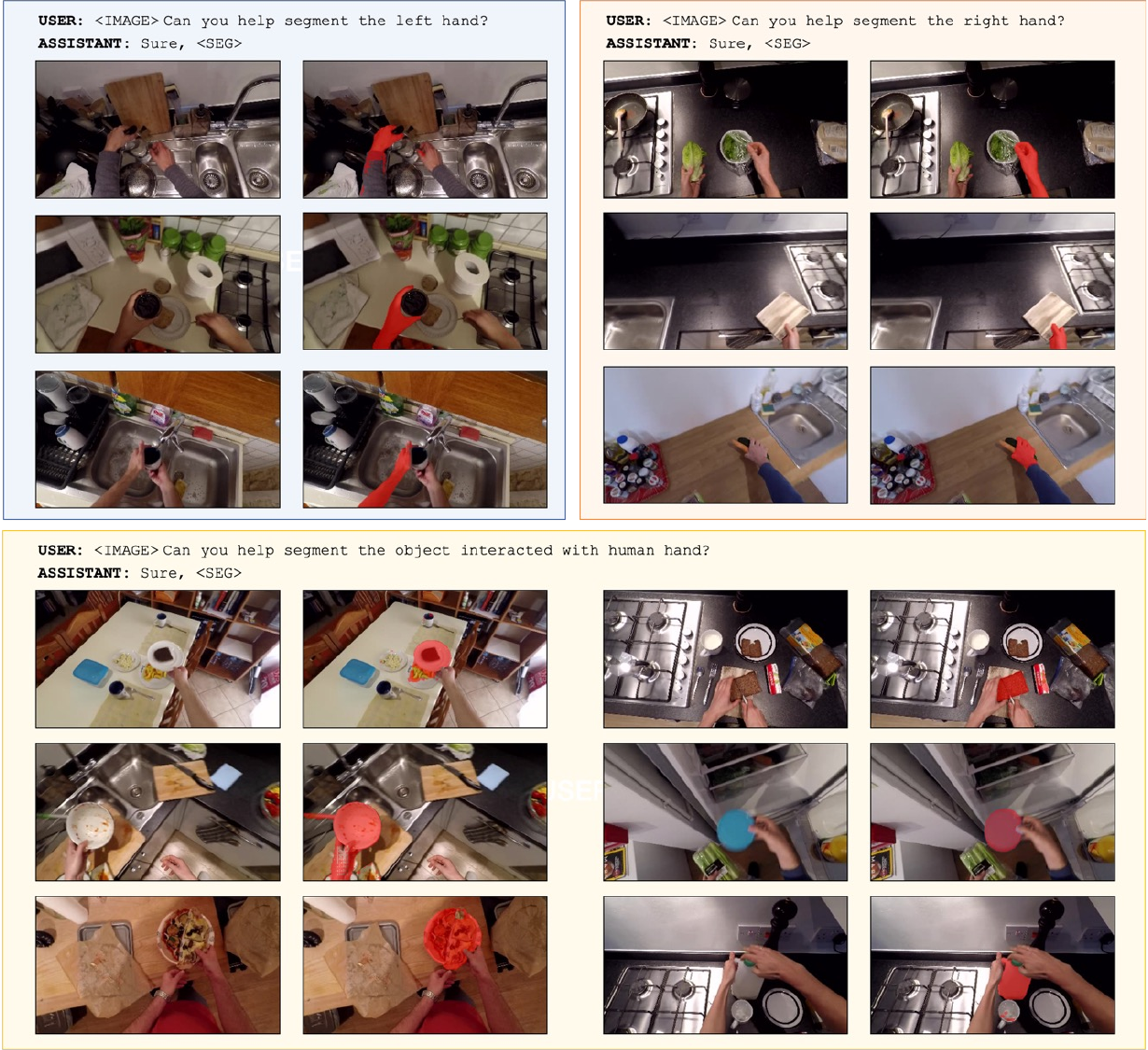}
\end{center}
\vspace{-0.5cm}
\caption{\textbf{The segmentation results by prompting LISA~\cite{lai2023lisa} with text prompts.} We prompt LISA with different text prompt, to get the segmentation of ``objects interacted with hands'' (red boxes), ``left hand'' (blue boxes) and ``right hand'' (red boxes). During the domain adaptation phase of \Ours, these segmentations could serve as supervision to fine-tune the class-agnostic object detector.}
\label{fig: supp_lisa}
\end{figure*}
}

\clearpage
\appendix
\setcounter{page}{1}
\maketitlesupplementary

In the supplementary material, we offer additional results and analysis of our work. Initially, in~\cref{sec: add}, additional demonstrations are provided, accompanied by detailed analysis. Following this, \cref{sec: sup} delves into discussions regarding the supervision complexity of the proposed \Ours method. Furthermore, ~\cref{sec: domain_gap} presents statistics that highlight the domain gaps between the \textit{Epic-Kitchens} and \textit{EGTEA} datasets. This data forms the basis for our experiments in the second scenario, which focuses on cross-dataset adaptation. \cref{sec: upper} elaborates the details for the implementation of baselines in our experiments. These details provide further insights into the methodologies and rationale behind our experimental setups.

\section{Additional Demonstrations} \label{sec: add}
Accompanying the main paper, we have included two video demonstrations to visually illustrate the efficacy of the proposed \Ours. We attach the supplementary material with a \texttt{mp4} video to show the demonstrations.

\subsection{Pour the Coffee.}
This demonstration is chosen from D1 test set in \textit{Epic-Kitchens} with the \texttt{uid} of 13838. The source detector is trained on D2. During the domain adaptation phase, we use frames in D1 train set to fine-tune the detector.

In this demonstration ``pour the coffee'', the limitations of the source detector become evident. It fails to correctly identify that the left hand is empty, instead consistently generating false positive predictions around the oven area. In contrast, the adapted detector consistently performs accurately, without such errors. Moreover, with the adapted detector, the predictions for the right hand's interaction with the coffee container remain remarkably stable, both temporally and spatially. This enhanced detection capability allows the action recognizer to accurately classify the action as ``pouring'' as opposed to more generic actions like ``get'' or ``place''. This example underscores the refined accuracy and reliability that \Ours brings to video domain adaptation tasks.

\subsection{Cut Lemon.}
This demonstration is chosen from D1 test set in \textit{Epic-Kitchens} with the \texttt{uid} of 14279. The source detector is trained on D3. In the "cut lemon" demonstration, the complexity of the action, involving two interactive objects: a knife and a lemon, poses a significant challenge. The source detector consistently fails to recognize the knife, resulting in a low confidence score for the accurate action label ``cut''. However, after adaptation, the detector shows marked improvement. It precisely localizes the knife in the right hand initially, then recognizes both the knife and lemon slices as a combined object during the action, and finally identifies the knife again as the action concludes. This enhanced object detection capability substantially improves the accuracy of the action recognition, effectively distinguishing between different phases of the ``cut'' action.

These demonstrations clearly illustrate how critical objects are in action recognition. With the adapted detector, the action recognizer receives more detailed object-level information. This refinement in the detection process significantly enhances the precision of action prediction, demonstrating the reciprocal benefits of improved object detection on the overall accuracy of action recognition.

\SuppLISA{!t}
\LISA{!t}
\SuppSAM{!t}

\section{Supervision Complexity Analysis} \label{sec: sup}

In this section, we analyze the supervision introduced during the domain adaptation phase of \Ours, contrasting it with a supervised baseline. The crux of \Ours lies in its strategy to bridge the domain gap by adapting the class-agnostic object detector with enriched object information from the target domain. To achieve this, we fine-tune the object detector using only a sparse set of frames annotated with object bounding boxes. With the advent and proliferation of large-scale multi-modal and foundational models, acquiring the necessary bounding boxes for \Ours has become increasingly feasible. In the following discussion, we explore two straightforward methods for obtaining these bounding boxes automatically.

\noindent \textbf{LISA~\cite{lai2023lisa}.} As a multi-modal LLM, LISA proposes \textit{reasoning segmentation} and is capable of producing segmentation masks based on the given text prompt. As shown in~\cref{fig: supp_lisa}, we ask/prompt LISA to segment the interacted object, left hand and right hand separately and get the segmentation results. Such process makes the supervison in the domain adaptation phase of \Ours totally free.

\noindent \textbf{SAM~\cite{kirillov2023segment}.} SAM an image segmentation model trained with billions of high-quality masks, supporting bounding boxes and points as prompts while demonstrating exceptional segmentation quality. As depicted in~\cref{fig: supp_sam}, we utilize SAM with point-based prompts to obtain segmentation results for objects interacted with hands, as well as for the left and right hands separately. In line with the methodology described in the main paper, such a process equates one bounding box annotation to a simple mouse click.

The segmentation masks generated automatically can be seamlessly converted into bounding boxes, which are then used in the fine-tuning of the class-agnostic object detector. By specifically focusing on interacted objects and hands in each frame, \Ours substantially reduces the need for extensive supervision. Moreover, with the aid of the above models, this supervision can be further automated, enhancing efficiency and reducing the labor-intensive aspects of the process.

We include the experiment results of the adaptation within \textit{Epic-Kitchens} dataset in~\cref{tab: automatic_label}, where we assess the performance of \Ours using automatically generated labels for segmentation. This segmentation is subsequently converted into bounding boxes, which are then utilized to fine-tune the class-agnostic object detector during the domain adaptation phase. In~\cref{tab: automatic_label}, ``LISA'' denotes the use of the LISA model to segment 100 frames for evaluation purposes. ``LISA+'' indicates an additional step where noisy labels are filtered out post-segmentation before evaluation. ``SAM + Point'' refers to the use of a point-based SAM model to generate the labels.

\quad

\quad

\section{Domain Gap} \label{sec: domain_gap}

In~\cref{sec: 4.3} of the main paper, our second scenario, ``adaptation across different datasets'', uses the combination of three kitchens in \textit{Epic-Kitchens} as the source domain and 7 recipes in \textit{EGTEA} as the target domains.

As different datasets, \textit{Epic-Kitchens} and \textit{EGTEA} exhibit domain gaps with respect to various aspects. We give details of these gaps, particularly focusing on the actions. \cref{supp: overlapping_ek_egtea} shows statistics that illustrate the overlap in action vocabulary between the two datasets, along with the average frame number for each video segment. For instance, \textit{Epic-Kitchens} encompasses 97 different action classes, whereas \textit{EGTEA} contains 19 action classes that are all overlapped with those in \textit{Epic-Kitchens}. This comparative analysis sheds light on the similarities and differences in action representation between the two datasets.

\begin{table}[!ht]
\begin{center}
\tablestyle{5.pt}{1.0}
\small
\begin{tabular}{c|cc}

Dataset             & \# Action   & \# Average Frames per Video Clip \\ \shline
\textit{\textit{Epic-Kitchens}} & 97       &179        \\
\textit{EGTEA}        & 19        &87            \\
\rowcolor{gray!10} \textbf{Overlap}      & \textbf{19/19}    & -           \\ 
\end{tabular} \vspace{-1.2em}
\end{center}
\caption{\textbf{Domain gap between \textit{Epic-Kitchens} and\textit{ EGTEA}.} }
\label{supp: overlapping_ek_egtea}
\end{table}

\section{Comparable Fully-supervised Approach} \label{sec: upper}

In the experiments section, we treat the fully-supervised method as the upper bound of adaptation performance. Here, "fully-supervised" implies the utilization of a "comparable" number of fully annotated video data for end-to-end fine-tuning of the action recognizer during the adaptation phase. The term "comparable" specifically refers to the number of video clips that correspond to the sparse set of frames employed in \Ours. This approach allows for a fair and relevant comparison of the performance between the fully-supervised method and \Ours.

We describe the mapping scheme utilized to correlate each frame employed in the fine-tuning of the detector with the corresponding video clips in the fully-supervised method. This mapping is achieved by associating each frame with the video clip it originates from in the raw dataset. However, there are edge cases to consider: (1) multiple frames from the frame fine-tuning set originate from the same video segment; and (2) instances where some frames do not correspond to any specific clip, typically being frames located ``between'' different video clips. For these edge cases, we change the frames to guarantee the one-to-one unique mapping between frames and video clips. Employing this mapping scheme enables us to establish a direct relationship between object-level and video-level supervision. This approach is instrumental in validating the efficiency of \Ours, not only from the perspective of performance but also in terms of resource consumption.



\end{document}